\pdfoutput=1

\documentclass[11pt]{article}

\usepackage{acl}
\usepackage{graphicx}
\usepackage{times}
\usepackage{latexsym}
\usepackage{multirow}
\usepackage{amsmath}

\usepackage[T1]{fontenc}

\usepackage[utf8]{inputenc}

\usepackage{microtype}

\title{BEIKE NLP at SemEval-2022 Task 4: Prompt-based Paragraph Classification for Patronizing and Condescending Language Detection}

\author{Yong Deng, Chenxiao Dou\thanks{\quad Corresponding Author}, Liangyu Chen \\
  \bf{Deqiang Miao, Xianghui Sun, Baochang Ma, Xiangang Li} \\
  AI Center@Beike, China \\
  \texttt{\{dengyong013, douchenxiao001,chenliangyu003\}@ke.com}
}

\begin{document}
\maketitle

\begin{abstract}
  PCL detection task is aimed at identifying and categorizing language that is patronizing or condescending towards vulnerable communities in the general media. 
  Compared to other NLP tasks of paragraph classification, the negative language presented in the PCL detection task is usually more implicit and subtle to be recognized, making the performance of common text-classification approaches disappointed. 
  Targeting the PCL detection problem in SemEval-2022 Task 4, in this paper, we give an introduction to our team's solution, which exploits the power of prompt-based learning on paragraph classification.
  We reformulate the task as an appropriate cloze prompt and use pre-trained Masked Language Models to fill the cloze slot.
  For the two subtasks, binary classification and multi-label classification, DeBERTa model is adopted and fine-tuned to predict masked label words of task-specific prompts.
  On the evaluation dataset, for binary classification, our approach achieves an F1-score of 0.6406; for multi-label classification, our approach achieves an macro-F1-score of 0.4689 and ranks first in the leaderboard.
  
\end{abstract}

\section{Introduction}

Patronizing and Condescending Language (PCL) towards vulnerable communities in the media has become a hot issue, which is heatedly discussed in society at present.
The language often refers to the discourse which is published with a mixture of pity and superiority towards unprivileged groups.
Such attitude is always adopted unconsciously or even out of kindness, however, its negative effect indeed exacerbates social prejudice and routinizes discrimination towards disadvantaged people~\cite{ng2007language}. 
Despite the PCL emerging in numerous media, its style can not be precisely captured by far, since the usage of PCL is commonly unintended, unaffected, and subjective.
This poses many challenges to the detection of PCL, which attracts great attention from the NLP community.
Even though there exist substantial NLP studies on paragraph identification in various fields, research specifically for PCL identification has not been yet seriously introduced~\cite{perezalmendros2020dont}.
Nowadays with social discrimination continuously rising, an effective approach, which can automatically identify PCL towards vulnerable communities, becomes more and more necessary and important to our society. 

To encourage more research on the PCL problem, SemEval-2022 Task 4~\cite{perezalmendros2022semeval} provides an English PCL dataset for language-modeling study and evaluation.
The main task contains two PCL-classification subtasks, one for binary classification (SubTask 1) and the other for multi-label classification (SubTask 2).
Given a paragraph, SubTask 1 is aimed to identify whether or not it is written with PCL style.
F1 over the positive class is taken as the evaluation metric of this task.
And for the same paragraph, the goal of SubTask 2 is to classify which specific PCL categories it belongs to. 
In detail, there are 7 different PCL types needed to be recognized, and one paragraph can have up to 7 PCL labels at the same time.
For evaluation, SubTask 2 introduces macro F1 over the 7 classes as the metric.

In this paper, we propose our approach targeting PCL detection, which is submitted to SemEval-2022 Task 4.
To tackle the sparsity and implicity issues of PCL language, we first reformulate the task as a specific form of cloze prompt, and then apply prompt-based learning on it to predict the appropriate label words and corresponding task labels.
To take advantage of the pre-trained Language Model (LM), DeBERTa~\cite{he2020deberta} is adopted as our based model, which improves BERT~\cite{devlin2018bert} by introducing two novel techniques: disentangled attention and decoding enhanced masking.
The same classification method and model architecture are used for both subtasks to train and predict.
Experiments are conducted to show the effectiveness of our approach.
At last, during the evaluation phase, our approach achieves an F1 score of 0.6406 on SubTask 1 (rank 4th on the leaderboard), and a macro-F1 score of 0.4689 on SubTask 2 (rank 1st on the leaderboard).

\section{Related Work}
Condescending and patronizing treatment is a controversial social problem, which attracts attention from researchers of various fields~\cite{margic2017communication, huckin2002critical}. 
To the NLP community, although extensive work on detecting different kinds of harmful language is presented, the targeted language styles are often explicit, aggressive, and flagrant, which can be perceived obviously.
Unlike the previously studied language, the style of PCL language can not be easily sensed, since the usage of PCL is commonly unintended, unaffected, and subjective, which makes it a challenging identification problem.
In recent years, more and more researchers~\cite{mendelsohn2020framework, sap2019social} start researching this emergent NLP topic, but the room for progress is still large. 

Paragraph classification is a basic and important NLP task, which has been continuously studied by industry and academia.
Applications based on paragraph classification are increasingly permeating our lives, such as spam filtering~\cite{kumar2020predictive} and text sentiment analysis~\cite{gao2019target}.
Traditional classification models, such as SVM~\cite{joachims1998text} and XGBoost~\cite{chen2016xgboost}, predicts the category of paragraph mainly based on the statistical features extracted from raw text data.
With the appearance of BERT~\cite{devlin2018bert}, many researchers~\cite{croce2020gan,jin2020bert} achieve success on paragraph classification, by utilizing the contextualized word vectors of pre-trained language models.

In recent years, with the development of prompt-based learning~\cite{schick2020s,schick2020exploiting,brown2020language}, researchers start reformulating the paragraph classification problem as a masked language modeling problem. 
Unlike the traditional models, prompt-based models first predict the label word of mask in a prompt and then map the label word to the label of category by a verbalizer.
Thanks to the power of pre-trained language models, prompt-based models~\cite{hu2021knowledgeable,gu2021ppt} demonstrate their strength in performing few-shot or even zero-shot learning on the scenarios of paragraph classification with few or no labeled data. 
The above studies have shown that prompt-based learning is highly effective to solve the problem of paragraph classification.
To study the effect of prompt-based learning on identifying paragraphs of PCL, in the rest of this paper, our modeling method and experimental analysis are introduced.

\section{Our Approach}
In this section, we present our approach to utilize prompt-based learning for PCL detection.
We first give the overall paradigm of our model and then introduce the ensemble strategy used in this task.

\subsection{Prompt-Based Model}\label{prompt}

\begin{figure*}
    \centering
    \includegraphics[width=1.0\textwidth]{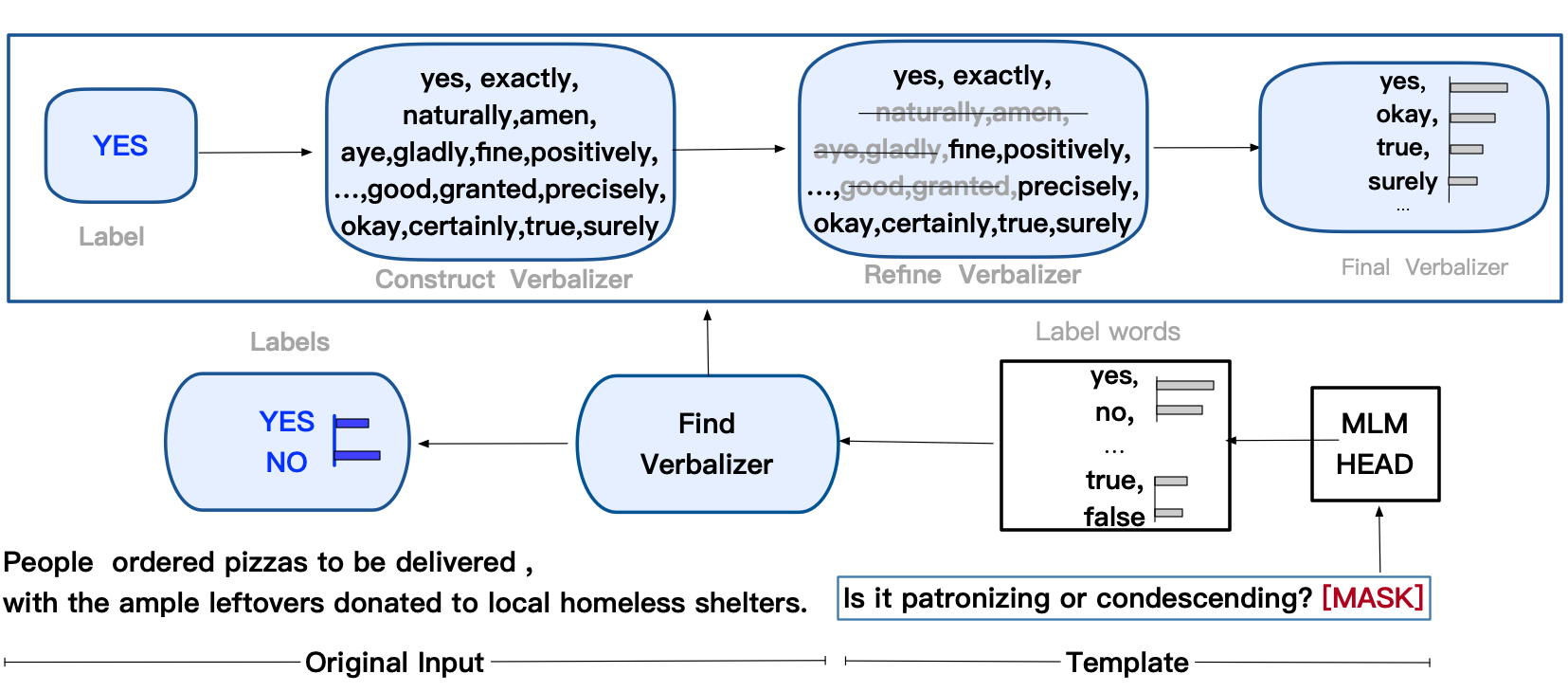}
    \caption{An example to illustrate our approach.}
    \label{fig:fig1}
\end{figure*}

Unlike other traditional classification methods which take original text as input, the prompt-based classification method first wraps the original text with some specific-designed template and then feeds the synthesized text into a language model for classification.
In this way, the paragraph classification problem is reformulated into a masked language modeling problem.
Formally, let $x = (x_0, x_1, ..., x_n)$ be a raw paragraph with a length of $n$ from the PCL dataset, $T$ be a prompt template and $x^T$ be the reformulated paragraph of $x$ using $T$.
For an example of SubTask 1 shown in Fig~\ref{fig:fig1}, with $x=$ \emph{``People  ordered pizzas to be delivered , with the ample leftovers donated to local homeless shelters.''} and $T$ = \emph{``Is it patronizing or condescending? [MASK]''}, we wrap them into $x^T$ as \emph{``People  ordered pizzas to be delivered , with the ample leftovers donated to local homeless shelters. Is it patronizing or condescending? [MASK]''}.

After the PCL problem is reformulated, an LM model $M$ can be used to compute the probability of each word $v$ in a vocabulary list to appear in the position of [MASK], denoted as $P({\rm[MASK]}=v|x^T)$. 
To answer the cloze question \emph{``Is it patronizing or condescending? [MASK]''}, if $M$ has a large probability to fill [MASK] with some words of the meaning related to \emph{``YES''}, then $x$ is predicted as PCL class; if words having similar semantics to \emph{``NO''} are more likely to be filled, then $x$ is deemed as non-PCL class.
In order to map the probabilities of predicted words to the probabilities of the labels \emph{``YES''} and \emph{``NO''}, we define a verbalizer to map a word $v$, from the label word set $V_y$, to a label $y$, form the label set $Y$.
To construct the verbalizer, we first use synonym dictionaries to find the synonyms of labels as the candidate label words, and then we select the top-k candidate words, according to their usage frequency in public corpora, as the final label words.
Assuming that all label words of the same label contribute equally to label prediction and the prior distribution of them is uniform, we use the mean of their predicted probabilities as the output probability of their corresponding label.
Then, given a reformulated paragraph $x^T$, its predicted label $\hat{y}$ can be obtained by
\begin{equation*}
  \hat{y} = \arg\max_{y \in Y} ( \frac{1}{|V_y|} \sum\limits_{v\in V_y} P({\rm [MASK]}=v|x^T) \notag
\end{equation*}
With the above definitions, in the training process, a cross-entropy loss is applied to fine-tune the model $M$ as our output model.

In SemEval-2022 Task 4, for the two subtasks, we use the same method to model and solve them.
The only difference is the template used for each subtask.
For the binary-classification subtask, only one template is provided to identify whether or not a paragraph is PCL class.
And for the multi-label-classification subtask, as there exist 7 PCL categories,  each of them is given with a unique template for type detection. 

\subsection{Ensemble Strategy}\label{ensemble}
To improve the robustness of the proposed approach, we take advantage of the cross-validation to make our model more stable and reliable. 
We randomly split the original PCL dataset into 10 parts, each of which has 1/10 samples of the dataset.
According to the hold-one-out strategy, 10 folds of the data can be acquired, and each fold has 9/10 samples for training and 1/10 samples for validation.
During the training process, the best-trained model for each fold is saved, and the average output probability of all models is taken as the final prediction score.
According to the score, each instance is assigned with a predicted label, and then the predicted labels are compared with the golden labels to compute the F1 score for validation.
The best models acquired in the validation process are kept for the final online evaluation.

\section{Experiment}

\begin{table*}
\centering
\begin{tabular}{lcccccc}
\hline
\multirow{2}{*}{\textbf{Strategy}} & \multicolumn{3}{c}{\textbf{SubTask 1}} & \multicolumn{3}{c}{\textbf{SubTask 2}} \\
  & R & P & F1 & macro-R & macro-P & macro-F1  \\
\hline
  CLS                                & 61.7\% & 61.0\% & 61.4\% & 41.7\% & 41.1\% & 41.3\%\\
  Prompt                             & 62.2\% & 61.8\% & 61.9\% & 44.3\% & 44.0\% & 44.4\%\\
  Prompt + Ensemble                  & 63.0\% & 61.6\% & 62.3\% & 46.0\% & 46.3\% & 46.4\%\\
  Prompt + Ensemble + R-Drop         & 63.1\% & 62.0\% & 62.5\% & 46.3\% & 46.4\% & 46.5\%\\
  Prompt + Ensemble + R-Drop + EDA   & 63.3\% & 63.0\% & 63.1\% & 46.5\% & 48.2\% & 47.5\%\\
\hline
\end{tabular}
\begin{tabular}{lc}
\hline
\end{tabular}
\caption{Experimental Results on SubTask 1 and SubTask 2}
\label{tab:tab1}
\end{table*}

\subsection{Settings}
The PCL dataset~\cite{perezalmendros2020dont} containing 10,469 paragraphs is shared by both SubTask 1 and SubTask 2.
The goal of SubTask 1 is to detect 993 PCL samples from the whole dataset, and the goal of SubTask 2 is to classify the 993 PCL samples into 7 specific PCL types.
In our experiments, all trials are performed with Nvidia Tesla A100 and large-DeBERTa is used as the based pre-trained language model.
Considering the small size of the task data, we take AdamW as the optimizer in the experiments with a learning rate of 1e-5 and a maximum epoch number of 10.
Early-Stop strategy is also used in our training process. 
For other parameters, we set the batch size as 16 and the maximum sequence length as 256.

\subsection{Used Strategies}
For the sake of result analysis, we list all strategies used in our approach towards SemEval-2022 Task 4 as the following.
\begin{itemize}
 \item \textbf{Prompt:} Prompt strategy is the method described in Section~\ref{prompt}.
\item \textbf{Ensemble:} Ensemble strategy is the method described in Section~\ref{ensemble}.
\item \textbf{CLS:} We directly feed the original text into DeBERTa model and apply a softmax layer on the CLS token for classification. This method is used as a comparison strategy, which does not reformulate the classification problem.
\item \textbf{EDA:} ~\cite{wei2019eda} is a method of data augment, which introduces 4 operations, Synonym Replacement, Random Insertion, Random Swap, and Random Deletion, to boost performance on paragraph classification.
\item \textbf{R-Drop:} ~\cite{wu2021r} is a method to regularize dropout, which minimizes the bidirectional KL-divergence between the distributions of two sub-models sampled by dropout, to reduce model randomness.
\end{itemize}

\subsection{Results and Analysis}
Table~\ref{tab:tab1} shows all strategy results with Precision, Recall, and F1.
From the results, the effect of each strategy we used can be clearly observed. 
The last method including all Prompt, Ensemble, EDA and R-Drop performs best compared to other settings.

To show the superiority of the Prompt paradigm, we first compare it with CLS, which uses the traditional paradigm of classification.
From the table, it can be seen that Prompt leads CLS by about 0.5\% F1 score in SubTask 1 and 3.1\% macro-F1 score in SubTask 2.
As Prompt has natural advantages on few-shot learning and the size of the PCL dataset is small, such result is not out of expectation, which also shows the superiority of Prompt on detecting PCL language.
With the limited training data, to avoid the occurrence of overfitting and improve the robustness of our model, we then apply Ensemble strategy to the learning process.
As shown in the table, with the robustness improved, a 0.4\% increase of F1 score is achieved in SubTask 1.
And for SubTask 2, the improvement is more significant with a 2\% increase of macro-F1 score, since the data size of each category in SubTask 2 is far less than 1/10 of that in SubTask 1. 
To further enhance model robustness, R-Drop is also adopted in our approach, which aims to reduce the randomness generated by the dropout module of network. 
Though the performance improvement of R-Drop is not big, the stability and convergence of our approach increases, making the results of repeated trials more similar.

Besides the techniques of modeling, we also try to improve the learning by enhancing the quality of the training data.
As described in the previous sections, the size of the training data is small and PCL language is often written in an implicit style. 
Therefore, EDA strategy is performed to augment the PCL data, which replaces implicit words with their synonyms, and transforms the structure of paragraphs variously, to produce more data for training. 
With the quality-improved data, our approach achieves a  0.6\% lift of F1 score on SubTask 1 and a 1\% lift of macro-F1 score on SubTask 2.

\section{Conclusion}
In this paper, we propose a prompt-based learning approach for PCL detection, based on pre-trained language models.
To reformulate the PCL detection problem into a masked language modeling problem, the detail of how our prompt template is designed have been fully discussed.
To improve the robustness of our model, we also introduce several other techniques, which have a positive impact on prompt-based learning.
Through experimental comparison, our approach is proven as an effective solution to detect PCL language.
As a result, in SemEval-2022 Task 4, our approach ranks 4th on the leaderboard of SubTask 1, and ranks 1st on the leaderboard of SubTask 2.

\bibliography{custom}

\end{document}